\documentclass[11pt,a4paper]{article}
\usepackage[hyperref]{acl2017}
\usepackage{times}
\usepackage{latexsym}

\usepackage{graphicx}
\usepackage[]{amsmath}
\usepackage{multirow}

\usepackage{url}

\aclfinalcopy 

\title{NE-Table: A Neural key-value table for Named Entities}

\author{
  Janarthanan Rajendran\thanks{\hspace{0.2cm}Equal Contribution}\\
  {\fontsize{11}{13} \selectfont University of Michigan}\\
  {\fontsize{11}{13} \selectfont \tt rjana@umich.edu}
  \\\And
  Jatin Ganhotra\footnotemark[1] \\
  {\fontsize{11}{13} \selectfont IBM Research}\\
  {\fontsize{11}{13} \selectfont \tt jatinganhotra@us.ibm.com}
  \\\AND
  Xiaoxiao Guo\\
  {\fontsize{11}{13} \selectfont IBM Research}\\
  {\fontsize{11}{13} \selectfont \tt xiaoxiao.guo@ibm.com}
  \\\And
  Mo Yu\\
  {\fontsize{11}{13} \selectfont IBM Research}\\
  {\fontsize{11}{13} \selectfont \tt yum@us.ibm.com}
  \\\And
  Satinder Singh \\
  {\fontsize{11}{13} \selectfont University of Michigan}\\
  {\fontsize{11}{13} \selectfont \tt baveja@umich.edu}
  \\\And
  Lazaros Polymenakos\thanks{\hspace{0.2cm}This work was done when the author was at IBM Research, NY.}\\
  {\fontsize{11}{13} \selectfont Amazon Alexa-AI Research}\\
  {\fontsize{11}{13} \selectfont \tt polyml@amazon.com}
  }

\date{}

\begin{document}
\maketitle
\begin{abstract}
Many Natural Language Processing (NLP)
tasks depend on using Named Entities (NEs) that are contained in texts and in external knowledge sources. 
While this is easy for humans, the present neural methods that rely on learned word embeddings may not perform well for these NLP tasks, especially in the presence of Out-Of-Vocabulary (OOV) or rare NEs. In this paper, we propose a 
solution for this problem, and present empirical evaluations on: a) a structured Question-Answering task, b) three related Goal-Oriented dialog tasks, and c) a Reading-Comprehension task\footnote{We create extended versions of dialog bAbI tasks 1,2 and 4 and OOV versions of the CBT test set - \url{https://github.com/IBM/ne-table-datasets/}}, which show that the proposed method can be effective in dealing with both in-vocabulary and OOV NEs. 
\end{abstract}


\section{Introduction}
\label{introduction}

We come across Named Entities (NEs) in many Natural Language Processing (NLP) tasks. 
In tasks such as Question-Answering (QA) and goal-oriented dialog, NEs play a crucial role in task completion. 
Examples include QA systems for retrieving information  
about courses offered at a university, and dialog systems that perform restaurant reservation, flight ticket booking, and so on. In many cases, these tasks also involve interaction with external knowledge sources such as DataBases (DB) which could have a large number of NEs. 
In these tasks NEs include people's names, restaurant names, locations etc. 

There has been a lot of interest in building neural methods for NLP tasks. Past work has developed multiple methods for addressing the unique challenges to neural methods posed by NEs. One straightforward method is to add each and every NE (including those in the DB) to the vocabulary. This method has been evaluated for only synthetic or small tasks \cite{neelakantan2015neural}. For real world tasks, especially those with large DBs, this causes an explosion in the vocabulary size and hence the number of parameters to learn. There is also the problem of not being able to learn \textit{good} neural embeddings for individual NEs, as individual NEs (e.g., a particular phone number) generally occur only a few times in a dataset. Another previously proposed method is to encode all the NEs with random embeddings and keep them fixed throughout \cite{yin2015neural}, but here we lose the meaning associated with the neural embeddings and risk 
interference and correlation with others in unexpected ways. 

A third method is to first recognize the NE-type with either NE taggers \cite{finkel2005incorporating} or entity linkers \cite{cucerzan:2007:EMNLP-CoNLL2007,guo-chang-kiciman:2013:NAACL-HLT},
and then replace them with NE-type tags. For example, all location names could be replaced with the tag \textit{NE\_location}. This prevents the explosion in vocabulary size; however, the system loses the ability to distinguish and reference different NEs of the same type. 
There is also the possibility of new NEs arising during the test time. In fact, many of the Out-Of-Vocabulary (OOV) words that arise during test time in many NLP tasks (e.g. \citeauthor{bordes2016learning-memN2N} \shortcite{bordes2016learning-memN2N}) are NEs. Furthermore, in many scenarios it is easier and accurate to work with the actual exact values of NEs rather than 
neural embeddings, like providing a phone number to a user or searching for a faculty name over a DB. None of the above neural methods have the ability to work with exact NE values.

In this paper, we propose a novel neural method that addresses all the aforementioned issues. There are three aspects to our method. 
\begin{itemize}
\itemsep-0.2em
    \item On-the-fly-generation: Neural embeddings for the NEs are generated on the fly using their context information. This avoids the explosion in vocabulary size, while still providing meaningful and distinguishable neural embeddings for the different NEs.
    \item Key-Value-Table: The generated embeddings are stored in a table (\textit{NE-Table}), with embeddings as the keys (key-embeddings) and exact NEs as the values (NE-values).
    \item On-the-fly-Retrieval: The NE-values can later be retrieved from the \textit{NE-Table} by attending over the key-embeddings, providing the ability to interact with exact NE values.
\end{itemize}

We demonstrate our method on 
a reading-comprehension task, a simple structured Question-Answering (QA) task, and three goal-oriented dialog tasks. 
Our method achieves 10\% increase in accuracy for Reading-Comprehension, 19\% increase for structured-QA and around 90\% increase for goal-oriented dialog tasks, with respect to their corresponding baselines.

\section{NE-Table: A Neural Key-Value Table for Named Entities}
\label{proposed_solution}
\begin{figure}[ht]
\centering
\includegraphics[width=0.45\textwidth]{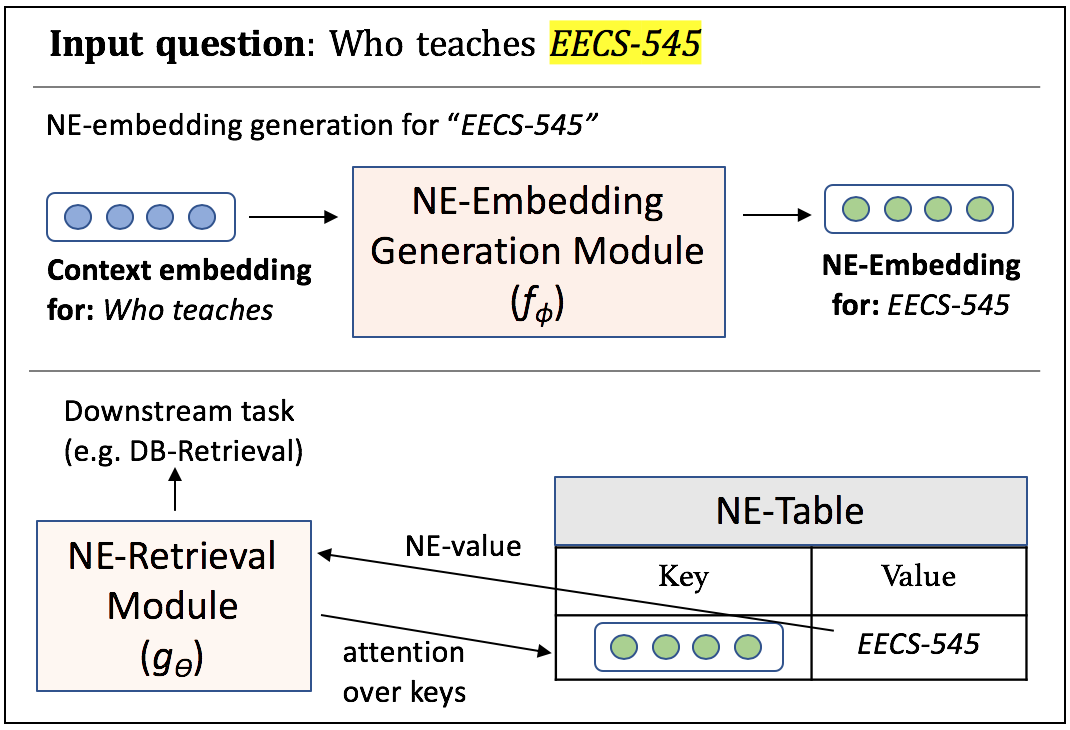}
\caption{For input question - \textit{Who teaches EECS-545}, the \textit{NE-Embedding Generation Module} ($f_\phi$) takes the context embedding as input and generates a NE-Embedding for the NE \textit{EECS-545}.  
The NE-Embedding is stored in \textit{NE-Table} with its actual value \textit{EECS-545}. The \textit{NE-Retrieval Module} ($g_\theta$) performs attention over the keys in \textit{NE-Table} to retrieve the NE-value. We show a simple example here to illustrate $f_\phi$ and $g_\theta$. Depending on the task, the context can vary and the \textit{NE-Table} can have more entries.} 
\label{fig_ne_table_idea}
\end{figure}

Our proposed method (Figure \ref{fig_ne_table_idea}) has three aspects.

\textit{On-the-fly-generation.} Neural embeddings for the NEs are generated on the fly using their context information (shown as the \textit{NE-Embedding Generation Module} in Fig \ref{fig_ne_table_idea}), instead of adding them to the vocabulary. The context information depends on the task. For a dialog task, the context is the full dialog so far, including the present utterance which has the NE in it. For the QA task, context is the sentence in which the NE appears. For the Reading Comprehension task, the sentence where the NE occurs or potentially the full story can be used as the context. The context could also include the NE-type information when available. The \textit{NE-Embedding Generation Module}, denoted ($f_\phi$), takes the context embedding as input and outputs the NE-Embedding. For our purposes, $f_\phi$ is an multi-layer perceptron (MLP). The problem of explosion in vocabulary size is avoided, as NEs are not part of the vocabulary and the NE-Embeddings are generated on the fly. Our proposed method also generates unique embeddings for different NEs with the same NE-type. This is better than replacing a NE with its NE-type as that results in all NEs with the same NE-type having the same embedding and hence, losing the ability to distinguish different NEs with the same NE-type. The generated NE-Embeddings are meaningful as they are learned from the context, in comparison to fixed random embeddings and can also be used as the learned neural embedding for that NE word from thereon.

\textit{Key-Value-Table.} As discussed in the previous section, there are many scenarios where it is easier and more accurate to work with the exact values of NEs rather than their neural embeddings, like providing a phone number to a user or searching for a faculty name over a DB. 
For this purpose, the generated NE-Embedding, along with its exact NE value is stored in a table, \textit{NE-Table}, as a key-value pair, with the embedding as key (key-embedding) and the exact NE as value (NE-value). 

\textit{On-the-fly-Retrieval.} The NE-value can later be retrieved from the \textit{NE-Table} by performing attention over the key-embeddings in the \textit{NE-Table}. This is performed by the \textit{NE-Retrieval Module} ($g_\theta$) shown in Figure~ \ref{fig_ne_table_idea}. For our purposes, $g_\theta$ is an MLP. The input to 
\textit{NE-Retrieval Module} also depends on the task. For 
dialog task, the dialog state vector is used, which has the information of the dialog so far. For 
QA task, the encoding of the input question is used. For 
Reading Comprehension task, the full story is used as input to the retrieval module. The retrieved NE-value can be used in the output utterance (e.g. providing a phone number) or to do an exact match over values in a DB (e.g. searching for a faculty name in a DB).

\begin{figure*}
\centering
\includegraphics[width=0.91\textwidth]{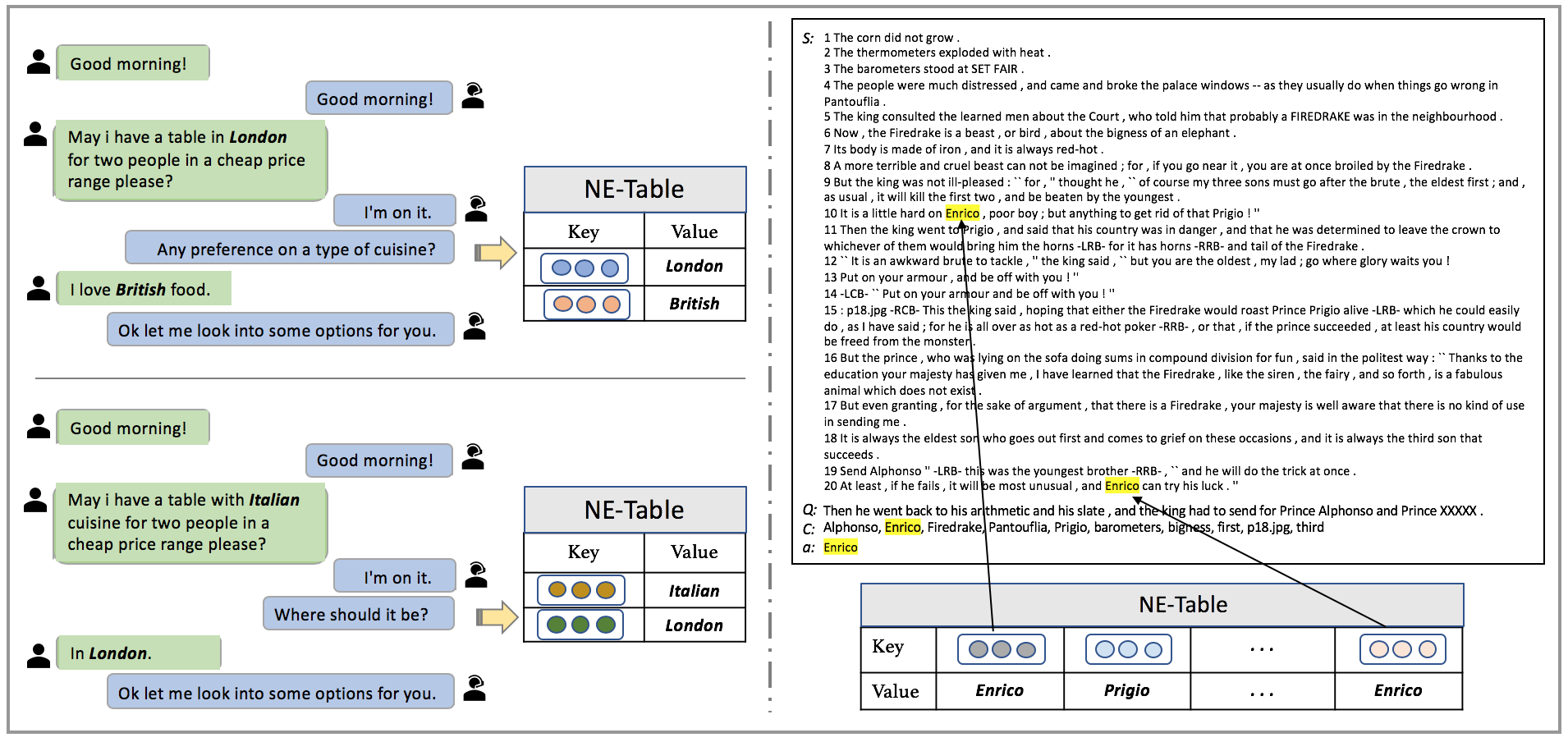}
\caption{\textit{Left:} \textbf{Two dialogs from bAbI task-1}. A user (in green) chats with a dialog system (in blue) to book a table. 
Each dialog has its own separate \textit{NE-Table} and a separate NE-Embedding is generated for the NE \textit{London} though it appears in both dialogs. \textit{Right:} \textbf{Question from CBT
}. NE \textit{Enrico} (highlighted in yellow) occurs twice in the context \textit{S}, where a separate NE-Embedding is generated for each occurrence.}
\label{fig_different_NE_Table}
\end{figure*}

While the matching of a NE-value retrieved from the \textit{NE-Table}, with other NEs in the DB is performed through exact value match, the actual retrieval of that NE from the \textit{NE-Table} happens using attention in the neural embedding space (using a dot product in our experiments). This allows the training of the \textit{NE-Retrieval Module} using the supervision obtained from the downstream module (e.g., a DB retrieval module) that uses the retrieved NE-value.
This also provides supervision for training the \textit{NE-Embedding Generation Module}. Our intuition is that, this would encourage the \textit{NE-Embedding Generation Module} to generate embeddings for the NEs such that the embeddings have relevant and enough information to allow the \textit{NE-Retrieval module} to attend and retrieve them correctly when required later.

Since the embeddings are generated on the fly using the context, the above method works equally well for new NEs that come during test time as it would for the NEs present in the training data. We show examples for \textit{NE-Table} for the dialog and Reading Comprehension task in Figure~\ref{fig_different_NE_Table}. A new, separate \textit{NE-Table} is created for each data instance based on the task. For example, in the dialog task, each dialog will have its own separate \textit{NE-Table}. Only the NEs that have appeared in the dialog so far will be present in its corresponding \textit{NE-Table}. The same NE occurring in different dialogs will have different dialog-context-dependent embeddings in their corresponding \textit{NE-Table}. Similarly, for the reading comprehension task, each story will have a separate \textit{NE-Table} with the NEs present in that story and for the QA task, each question will have a separate \textit{NE-Table}. Note that, a NE that occurs multiple times in the same dialog/story/question will also have multiple unique embeddings in the \textit{NE-Table} because of differing contexts as shown in Figure~\ref{fig_different_NE_Table} (right).

\section{Experiments and Results}
\label{experiments_and_results}

We evaluate our proposed method on three types of tasks: a reading-comprehension task, a structured-QA task and three goal-oriented dialog tasks. Our proposed method is generic and can be added to the state-of-the-art approaches for these tasks. But instead of implementing 3 separate specialized neural architectures, we chose the end-to-end memory network architecture from \citeauthor{sukhbaatar2015end} \shortcite{sukhbaatar2015end} as the base architecture for our tasks. This allows us to evaluate the advantage gained by adding our method to the base architecture instead of trying to get state-of-the-art performance in a particular task/dataset.

\subsection{Reading Comprehension Task}
The Children's Book Test dataset (CBT), built from children's books from ProjectGutenberg, was introduced by \citeauthor{hill2015goldilocks} \shortcite{hill2015goldilocks} to test the role of memory and context in language processing and understanding. 
Questions are formed by enumerating 21 consecutive sentences, where the first 20 sentences form the \textit{story}
(\textit{S}), and a \textit{word} (\textit{a}) is removed from the 21st sentence,  which then implicitly becomes the \textit{query} (\textit{q}). The specific task is to predict the correct answer word (\textit{a}) from a set of 10 candidate words (\textit{C}) present in the story or the query. 
We test our method on the NE questions subset of the CBT dataset.

We use the Window memory architecture proposed by \citeauthor{hill2015goldilocks} \shortcite{hill2015goldilocks} for the CBT dataset as our baseline. In Memory Networks \cite{sukhbaatar2015end}, each complete sentence of \textit{S} is encoded and represented in a separate memory slot. For the CBT, this setting would yield exactly 20 memories for \textit{S}. In Window memory, instead of a full sentence from the story, a phrase is encoded and represented in a separate memory slot. Each phrase $s$ corresponds to a window of text from the story \textit{S} centred on an individual mention of a candidate \textit{c} in \textit{S}. The window is constructed as span of words {$w_{i-(b-1)/2}$ ... $w_i$ ... $w_{i+(b-1)/2}$} where $b$ is window size and $w_i$ $\in$ \textit{C} is an instance of one of the candidate words in the question.
We perform two baseline evaluations: encoding the windows using a) Bag-of-Words (BoW) and b) LSTM \cite{hochreiter1997long}.

\begin{table}[t]
\begin{tabular}{ p{0.5\linewidth}|p{0.19\linewidth}|p{0.15\linewidth} }
Model & Validation & Test \\ \hline
W/O-NE-Table (BoW)	 &   49.55  & 41.69 \\ \hline
W/O-NE-Table (LSTM)  &   49.40	& 41.10 \\ \hline
With-NE-Table (BoW)  &	 57.05	& 51.28 \\ \hline
With-NE-Table (LSTM) &   55.75	& 51.08 \\ \hline
\end{tabular}
\caption{Results (accuracy \%) on CBT-NE dataset}
\label{table:cbt}
\end{table}

For each NE\footnote{The NEs present in the story are identified by the Stanford Core NLP NER system \cite{manning2014stanford}.}, the corresponding window is fed to an LSTM to create the context embedding. The context embedding is used as input to \textit{NE-Embedding Generation Module} ($f_\phi$), as shown in Figure~\ref{fig_ne_table_idea}, to generate the corresponding NE-Embedding, which is added to the \textit{NE-Table}. The NE-Embeddings are also added to window memory, in place of the NEs. The query (\textit{q}) embedding is used to attend over the memory (list of encoded window memory slots) to get relevant information from the memory. 
The internal state generated is given as input to the \textit{NE-Retrieval Module} ($g_\theta$), for retrieving the correct NE answer (\textit{a}). Table \ref{table:cbt} shows that replacing the baseline with our method achieves higher performance on both BoW and LSTM baseline models, across both validation and test sets. We use a window size of 5 as in \citeauthor{hill2015goldilocks} \shortcite{hill2015goldilocks}. We think that since the window size is small, both BoW and LSTM models perform similarly. We provide model training and hyperparameter details in the Appendix.

\begin{figure}[ht]
\centering
\includegraphics[width=0.40\textwidth]{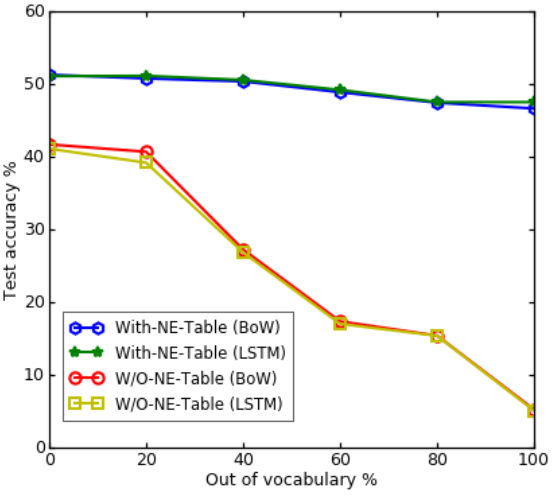}
\caption{Results on CBT-NE OOV test sets}
\label{fig_cbt_oov}
\end{figure}

To further evaluate the impact of OOV NEs, we created additional OOV test sets
by replacing 
NEs in the test set with new NEs not present in the train and validation sets. We generate 5 such OOV test sets with varying percentage of OOV NEs (20\%, 40\%, 60\%, 80\% and 100\%). Figure \ref{fig_cbt_oov} shows the comparison of our model with the baselines on OOV test-sets. The baseline models perform poorly as OOV\% increases, decreasing to as low as 5\% from 41\%. 
We observe only a slight reduction in accuracy for the NE-Table models from 51\% to 46\% because the new entities are also part of the windows, used to generate NE-Embeddings. These 
experiments illustrate that our model performance is robust to OOV NEs.

The next two tasks, structured-QA and goal-oriented dialog involve retrieval from an external DB. This is performed by the \textit{DB-Retrieval Module} ($h_\psi$), which uses a multiple-attention based neural retrieval mechanism. We describe this next and then present results on the 2 tasks.

\subsection{Multiple-attention based neural retrieval mechanism}
In both structured-QA and goal-oriented dialog tasks, the external information is present in a single database table, where each row corresponds to a new entity of interest and the columns correspond to the different attributes associated with it. For example, in our structured-QA (which is about course offerings at a university) DB, each row corresponds to a course and the columns correspond to course attributes, such as course number, course name, instructor name, etc. Each column of the table has a column heading, which labels the attribute of that column. These headings are also part of the vocabulary. While the non-NEs present in the DB are part of the vocabulary and represented by their learned neural embeddings, the NEs are not part of the vocabulary and are represented by their exact values.

The \textit{DB-Retrieval Module} performs attention over both attributes(columns) as well as rows to select the final cell(s) in 3 steps. In step 1, the column(s) that the final cell(s) belong to are selected by attention over the column heading embeddings. For the question \textit{Who teaches EECS545?}, step 1 selects the column \textit{'instructor name'}. In step 2, separate attention is performed over the column headings to select the columns, which are used to represent the rows (to retrieve the final cell) and column \textit{'course number'} is selected. Step 3 is to do attention over the rows. For each non-NE column selected in step 2, the cell embeddings are added together along each row, to generate an embedding for each row. Attention is performed over these row embeddings to select row(s). For each NE-column selected in step 2, a NE-value is retrieved from the \textit{NE-Table} to do an exact match search over that NE-column to select matching row(s). The intersection of these matching row(s) gives the final set of selected row(s), and their intersection with the set of column(s) selected in step 1 gives the retrieved cell(s). For our example, only one column is selected to represent the rows: \textit{'course number'}, which is a NE-column. Therefore, a NE value is retrieved from the \textit{NE-Table} (EECS545) and an exact match search is done over the \textit{'course number'} column.

The input to the \textit{DB-Retrieval Module} depends on the task. For the dialog task, the dialog state vector is used, which has the information of the dialog so far. For the QA task, the encoding of the input question is used. All the attention operations in our experiments are performed through dot product followed by a sigmoid operation, which allows for multiple selections. Additional details and further explanation of the retrieval mechanism with examples are provided in Appendix. Note that \textit{NE-Table} can potentially be used with other neural retrieval mechanisms. The multiple-attention mechanism described above is only one of the several neural retrieval mechanisms \cite{yin2015neural}.

\subsection{Structured-QA from DB}
The task here is to retrieve an answer (single cell in a table) from DB in response to structured one line questions. We used the details of course offerings at a university to create structured Question-Answer (QA) pairs. The DB is a single table of 100 rows (Courses) and 4 columns 
(Course Number, Course Name, Department, Credits)\footnote{number of unique course numbers - 100, unique course names - 96, unique dept names - 10 and unique credits - 4}, where course numbers and course names are treated as NEs. The QA pairs are generated 
through random sampling from the DB, following the format -\\
{\footnotesize
\texttt{\textbf{Q}: Col-1-type Col-1-value Col-2-type ?}\\
\texttt{\textbf{A}: Col-2-value\\}}
with the following being a specific example: \\
{\footnotesize \texttt{\textbf{Q}: Course Number EECS545 Credits ?
\textbf{A}: 4}.}

500 
QA pairs were created and
split randomly between training and test set (400-100), where the random split results in some NEs (OOV) in the test set, not present in the training set. The task was specifically constructed to be simple to show the impact of OOV NEs on the model performance and evaluate our proposed method.

The experiments were performed with two models. Both models use a 
Recurrent Neural Network (RNN) to encode the question and use the multiple-attention based neural retrieval mechanism 
to retrieve answers. The baseline model (\textit{W/O-NE-Table}) does not distinguish NEs from normal words, and all words (including NEs) that occur in questions and DB are part of the vocabulary. The \textit{With-NE-Table} model uses our proposed method and builds \textit{NE-table} (course numbers and course names are the NEs in this task). 
In the \textit{With-NE-Table} model, when a NE word is encountered, the hidden state of the RNN at the previous time step (word) is used as input to the \textit{NE-Embedding Generation Module} ($f_{\phi}$). The NE-Embedding generated by $f_\phi$ is then fed back 
to the RNN 
to continue encoding the question. The generated NE-Embedding is also stored in the \textit{NE-Table} associated with this question. The final hidden state of the RNN obtained after encoding the full question is provided as input to the \textit{DB-Retrieval Module} ($h_\psi$).

For our example, both models perform attention over the column headings to identify the correct column \textit{Credits} required for the answer. Then, both models attend over column headings to find the column \textit{Course Number} used for representing the rows. For \textit{W/O-NE-Table} model, since all course numbers are part of vocabulary, each row is represented by neural embeddings associated with course numbers and attention is done over the row embeddings. For \textit{With-NE-Table} model, since course numbers are NEs, each row is represented with exact course number values. A neural attention over \textit{NE-Table} is performed to return the NE value, \textit{EECS545}, which is then used to perform an exact match with the course number values. We provide model training details in Appendix.

\begin{table}[t]
\centering
\begin{tabular}{c|c}
Model & Retrieval accuracy (\%) \\\hline
W/O-NE-Table & 81.0 \\ \hline
With-NE-Table & 100.0 \\ \hline
\end{tabular}
\caption{Results on structured-QA task}
\label{table_qa_task}
\end{table}

Table \ref{table_qa_task} shows the retrieval accuracy for both models. While the test accuracy for \textit{With-NE-Table} is 100\%, it drops to 81\% for \textit{W/O-NE-Table} model. Further analysis shows that out of the 19\% drop, 11\% is due to OOV NEs encountered at test time. These OOV NEs are in the DB, and hence are part of the vocabulary for the \textit{W/O-NE-Table} model, but have random embeddings which did not change during the training time (as they were never encountered during the training). The rest 8\% drop can be attributed to the model's inability to learn good embeddings for unique NEs that were rarely seen during training. However, these issues do not pose a problem for our \textit{With-NE-Table} model, since we generate embedding for a NE on the fly for each question based on the context. This solves both problems: a) whether an NE occurred rarely or b) it was not present in training data at all. The \textit{With-NE-Table} model should also easily scale to large datasets with any number of NEs without drop in performance.  

\subsection{Goal-Oriented Dialog Tasks}
The Dialog bAbI tasks dataset was introduced by \citeauthor{bordes2016learning-memN2N} \shortcite{bordes2016learning-memN2N} as a testbed to break down the strengths and shortcomings of end-to-end 
goal-oriented dialog systems
The task domain is restaurant reservation and there are 5 tasks - Task 1: Issuing API calls, Task 2: Updating API calls, Task 3: Displaying Options, Task 4: Providing extra information and Task 5: Conducting full dialogs (combination of tasks 1-4). The system is evaluated in a retrieval setting. At each turn of the dialog, the system has to select the correct response from a list of possible candidates.

In the original bAbI tasks, DB-Retrieval is bypassed by providing all possible system utterances with all combinations of information \textit{pre}-retrieved from the DB in a large candidate response list. We extend the original testbed and propose a new testbed, which is closer to real-world restaurant reservation, by adding an actual external DB so that the system can also be tested on the ability 
to retrieve the required information from the DB. We evaluate our method on extended versions of task 1,2 and 4\footnote{Task 3 requires to learn to sort. \citeauthor{bordes2016learning-memN2N} \shortcite{bordes2016learning-memN2N} achieve close to 0\% accuracy on it.
Therefore, we decided to skip tasks 3 and 5 (task 5 includes task 3 dialogs) to focus on evaluating our proposed method.}.

For our experiments, we use an end-to-end memory network similar to \citeauthor{bordes2016learning-memN2N} \shortcite{bordes2016learning-memN2N}, except that we encode sentences using an RNN, while they use BoW encoding. The encoded sentences, which are part of the dialog history, are stored in the memory and the query (last user utterance) embedding is used to attend over the memory to get relevant information from the memory. 
The generated internal state is used to select the candidate response, and is also given as input to the \textit{DB-Retrieval Module} ($h_\psi$). The DB is used to identify the NEs along with their types (if a word is present in a NE-column in the DB it is a NE; the column where it appears gives its NE-type).\footnote{This simple method (based on exact match) though works for this dataset, is not very effective, as plural or abbreviated NEs will not match.}

The experiments are performed on two models:
\begin{itemize}
\itemsep-0.2em
    \item \textit{W/O-NE-Table} model (the baseline model) - All input words including NEs are part of the vocabulary. For NEs, however, their embedding given to the sentence encoder RNN is the sum of the NE word embedding and the embedding associated with its NE-type.
    \item \textit{With-NE-Table} model (uses our proposed method) - When an NE is encountered in the dialog, the last hidden state of the RNN encoding the sentence is used as input to the \textit{NE-Embedding Generation Module} ($f_\phi$). The NE-Embedding generated is stored in the \textit{NE-Table}. The generated NE-Embedding and the embedding associated with its NE-type are fed to the RNN. 
\end{itemize}
Note that both the models have access to the information whether a given word is a NE and its NE-type. Supervision is provided for candidate response selection and all attention operations performed during \textit{DB-Retrieval}, for both models.


\subsubsection{Extended dialog bAbI tasks 1 and 2}
In the original bAbI task 1, the conversation between the system and the user involves getting information necessary to issue an \textit{api\_call}. In task 2, the user can ask the system to update his/her preferences (cuisine, location etc.). The system has to take this into account and make an updated \textit{api\_call}. In our extended version, once the system determines that the next utterance is an \textit{api\_call}, the system also has to actually retrieve the restaurant details from the DB (rows) which match user preferences. The system is evaluated on having conversation with the user, issuing \textit{api\_call} and retrieving the correct information from 
DB. 
The DB is represented as a single table, where each row corresponds to a unique restaurant and columns correspond to attributes, e.g. cuisine, location etc.

Both \textit{W/O-NE-Table} and \textit{With-NE-Table} models, first select the four relevant (cuisine, location, price range and number of people) columns 
to represent each row (restaurant). The \textit{W/O-NE-Table} model then selects the rows using attention over the row embeddings obtained through the combined (additive) representation of the four selected attributes. The \textit{With-NE-Table} model splits the row selection into two simpler problems. For cuisine and location (which are NEs), one NE value each is retrieved from the \textit{NE-Table} and an exact match is performed in the DB. The neural embeddings of the non-NE attributes (price range and number of people) are added to perform attention for selecting rows. The final retrieved rows are the intersection of the rows selected by NE column and non-NE column based selections.

\begin{table}[ht]
{\fontsize{9}{11}\selectfont
\begin{tabular}{@{}l|p{1.4cm}|p{1.4cm}|p{1.75cm}@{}}
\multirow{2}{*}{Model} & DB-Retrieval & \multirow{2}{*}{Per-Dialog} & Per-Dialog + DB-Retrieval \\ \hline
\multicolumn{1}{@{}p{1.6cm}|}{Task 1} & & & \\
W/O-NE-Table & 10.2 (7.0) & 100 (90.3) & 10.2 (6.7)\\
With-NE-Table & 98.5 (99.0) & 98.8 (99.0) & \textbf{97.3 (98.0)} \\ \hline
\multicolumn{1}{@{}p{1.6cm}|}{Task 2} & & & \\
W/O-NE-Table & 0.8 (1.0) & 100 (100) & 0.0 (0.1)\\
With-NE-Table & 99.6 (99.8) & 100 (99.9) & \textbf{99.2 (99.7)} \\ \hline
\multicolumn{1}{@{}p{1.6cm}|}{Task 4} & & & \\
W/O-NE-Table & 0.0 (0.0) & 100 (100) & 0.0 (0.0)\\
With-NE-Table & 100 (100) & 100 (100) & \textbf{100 (100)} \\ \hline 
\end{tabular}
\caption{Results for extended bAbI tasks 1, 2 and 4. \% accuracy for Test and Test-OOV (given in parenthesis). \textit{DB-Retrieval} : Retrieval accuracy for rows (task 1,2 - \textit{all} restaurants matching user preferences) and a particular cell (task 4 - restaurant phone number/address). \textit{Per-Dialog} : Percentage of dialogs where every dialog response is correct.
Training details and hyperparameter values are provided in Appendix.}
\label{table_babi_task1_2_4}
}
\end{table}

The results for tasks 1 and 2 are shown in Table \ref{table_babi_task1_2_4}. The \textit{With-NE-Table} model achieves close to 100\% accuracy in both tasks, while \textit{W/O-NE-Table} performs poorly. During DB retrieval, for the \textit{With-NE-Table} model, 
two NEs are chosen from the \textit{NE-Table} and exact matching is done over different cuisines and locations in the DB,
but embeddings for these NEs are learned for \textit{W/O-NE-Table}. This results in poor DB-Retrieval 
for \textit{W/O-NE-Table} for less frequent/ OOV location/cuisine values. Both models perform well in \textit{Per-dialog} accuracy as it does not involve DB retrieval\footnote{The system responses in tasks 1/2/4 do not contain any NEs, but the system still needs to understand user utterances which might have NEs.}.
The Per-Dialog accuracy is high for both models on the normal test set. However, for task 1 OOV-test set, \textit{W/O-NE-Table} model is affected by OOV-NEs (90.3\%), while \textit{With-NE-Table} model performance is robust (99.0\%).

\subsubsection{Extended dialog bAbI task 4}
The original task 4 starts at the point where a user has decided a particular restaurant. The system is given information (location, phone number, address etc.) about \emph{only} that restaurant as part of the dialog history and the user can ask for its phone number, address or both. For a given user request e.g. address, the task is to select the correct response with the restaurant's address from a list of candidate responses. These candidate responses have phone number and address information for all the restaurants mentioned in the DB.

In our extended version, even though the user has decided a particular restaurant, its corresponding information is not provided as part of dialog history. This makes the task harder but more realistic. Now, the system needs to search for phone number/address for the restaurant from the full DB while in the original task, the phone number/address is already provided as part of dialog history. In the extended version, the NEs in candidate responses are replaced with their NE-type tags. For example, \textit{Suvai\_phone} is replaced with \textit{NE\_phone}. The system has to select the candidate with correct NE-type tag and then replace the tag with the actual NE-value retrieved from the DB, similar to \citeauthor{williams2017hybrid} \shortcite{williams2017hybrid}. This setting is closer to how a human agent would do this task.

For \textit{With-NE-Table} model, the restaurant name that appears in the dialog would be stored in \textit{NE-Table}. When the user asks for information such as phone number, the restaurant name stored in \textit{NE-Table} is selected and used for retrieving its 
phone number from the DB. In \textit{W/O-NE-Table} model, all input words (including NEs) are part of vocabulary and phone number is selected by neural embedding attention over all restaurants names. 

The results for task 4 are shown in Table \ref{table_babi_task1_2_4}. We observe that both models perform well in Per-dialog accuracy. The \textit{W/O-NE-Table} model fails in DB-retrieval (0\%) because it needs to learn neural embeddings for all restaurant names, while \textit{With-NE-Table} performs well (100\%) as it uses our proposed method to generate NE-Embeddings on the fly 
and use the actual NE values later for exact value matching over restaurant names in the DB.

\section{Related Work}
\label{related_works}
\textbf{NE in QA}: 
\citeauthor{neelakantan2015neural} \shortcite{neelakantan2015neural} and \citeauthor{yin2015neural} \shortcite{yin2015neural} transform a natural language 
query to a program that could run on DBs, but those approaches are only verified on small or synthetic 
DBs.
Other papers dealing with large Knowledge Bases (KB) usually rely on entity linking techniques
\cite{cucerzan:2007:EMNLP-CoNLL2007,guo-chang-kiciman:2013:NAACL-HLT}, which links entity mentions in texts to KB queries. 
Recently, \citeauthor{liang2016neural} \shortcite{liang2016neural} extended end-to-end neural methods to QA over KB, which could work for large KB and large number of NEs. However, their method still relies on entity linking 
to generate a short list of entities linked with text spans in the questions, in advance. \citeauthor{yin2015neural} \shortcite{yin2015neural} propose 'Neural Enquirer', a neural network architecture similar to the neural retrieval mechanism used in this work, to execute natural language queries on DB. They keep the randomly initialized embeddings of the NEs fixed as a method to handle NEs and OOV words.

\textbf{NE in Dialog}: There has been a lot of interest in end-to-end training of dialog systems \cite{vinyals2015conversationalmodel,serban2016building,lowe2015ubuntu,kadlec2015ubuntu,shang2015responding,guo2016learning}.
Among recent work, \citeauthor{WilliamsZweig16} \shortcite{WilliamsZweig16} use an LSTM model that learns to interact with APIs on behalf of the user; \citeauthor{dhingra-EtAl:2017:Long1} \shortcite{dhingra-EtAl:2017:Long1} use reinforcement learning to build the KB look-up in task-oriented dialog systems. But the look-up actions are defined over each entity in the KB and is therefore hard to scale up. Most of these papers actually do not discuss the issue of handling NEs though they are present. \citeauthor{williams2017hybrid} \shortcite{williams2017hybrid} propose Hybrid Code Networks and achieve state-of-the-art on Facebook bAbI dataset, but approach involves a developer writing domain-specific software components.

\textbf{NE in Reading Comprehension and others}:
For certain tasks such as Machine Translation and summarization, neural copying mechanisms \cite{gulcehre2016pointing,GuLLL16} have been proposed 
for handling OOV words. 
Our \textit{NE-Table} method can be used along with such copying mechanisms for cases like dialog generation.

\section{Conclusion and Future work}
\label{conclusion}
In this paper, we proposed a novel method for handling NEs in neural settings for NLP tasks. Our experiments on the CBT dataset illustrate that the models with \textit{NE-Table} perform better than models without \textit{NE-Table}, and clearly outperform the baseline models on the OOV test sets. We observe similar results for our experiments on the structured-QA task and goal-oriented bAbI dialog tasks. We also show that our method can be used for NEs in the external DB provided. 
Overall, these experiments show that the proposed method can be useful for various NLP tasks where it is beneficial to work with actual NE values, and/or it is hard to learn good neural embeddings for NEs. 

In future, we are interested in testing the proposed method with retrieval mechanisms such as 'Neural Enquirer' \cite{yin2015neural}, which can work with multiple tables. We are also interested in exploring the use of pre-trained embeddings: word2vec \cite{mikolov2013distributed},
ELMo \cite{peters2018deep} etc., to bootstrap our learned NE-embeddings. We are also interested in evaluating our proposed method on tasks that are more unstructured and requires more free-form generation, e.g. machine translation and dialog generation.


\bibliography{acl2017}
\bibliographystyle{acl_natbib}

\newpage

\appendix

\section{Model Training and Hyperparameter Details}
\label{appendix_tr}

\subsection{Reading Comprehension - CBT}
The hyperparameters used for baseline \textit{W/O-NE-Table} models are as follows: hops = 1, embedding\_size = 100, batch\_size = 16. The \textit{W/O-NE-Table (LSTM)} model uses an LSTM for encoding windows with hidden\_units = 100. The entire model is trained using stochastic gradient descent (learning rate = 0.05), minimizing a standard cross-entropy loss between predicted answer $\hat{a}$ and the correct answer $a$. We use the same embedding matrix for encoding both story and the query.

For our \textit{With-NE-Table} models, we use the same hyper-parameters as mentioned above.
The \textit{With-NE-Table (LSTM)} model uses separate LSTM cells for a) generating the context embedding for \textit{NE-Embedding Generation Module} ($f_\phi$) and b) encoding the windows.

\subsection{Structured-QA}
Both the \textit{With-NE-Table} and  \textit{W/O-NE-Table} models use the following hyperparameters : embedding\_size = 20, batch\_size = 16. The RNN used for encoding the sentences has hidden\_units = 20. The model is trained using Adam (learning rate = 0.01, epsilon = 1e-8). Supervision is provided for DB-Retrieval attentions and standard cross-entropy loss is used.

\subsection{bAbI dialog tasks}
Both the \textit{With-NE-Table} and  \textit{W/O-NE-Table} models use the following hyperparameters : hops = 3, embedding\_size = 40, batch\_size = 16. The RNN used for encoding the sentences has hidden\_units = 40. The model is trained using Adam (learning rate = 0.001, epsilon = 1e-8). Supervision is provided for both the candidate response selection and DB-Retrieval attentions and standard cross-entropy loss is used.

\section{Extended dialog bAbI tasks' examples}
\label{appendix_a}

\subsection{Original dialog bAbI task 1:}
U: Hi !\\
\texttt{S: Hello, what can I help you with today?}\\
U: I'd like to book a table for six people in an expensive price range with British food.\\
\texttt{S: I'm on it.}\\
\texttt{S: Where should it be?}\\
U: In London.\\
\texttt{S: Ok, let me look into some options for you.}\\
\texttt{S: api\_call British, London, Six, Expensive}

\subsection{Extended dialog bAbI task 1:}
U: Hi !\\
\texttt{S: Hello, what can I help you with today?}\\
U: I'd like to book a table for six people in an expensive price range with British food.\\
\texttt{S: I'm on it.}\\
\texttt{S: Where should it be?}\\
U: In London.\\
\texttt{S: Ok, let me look into some options for you.}\\
\texttt{S: api\_call}\\
\texttt{S (Retrieved Information):}\\
\texttt{\{The\_Place, british, london, \dots\}}\\
\texttt{\{The\_Fancy\_Pub, british, london, \ldots\}}

\subsection{Original dialog bAbI task 2:}
U: Hi !\\
\texttt{S: Hello, what can I help you with today?}\\
U: I'd like to book a table for six people in an expensive price range with British food in London.\\
\texttt{S: Ok, let me look into some options for you.}\\
\texttt{S: api\_call British, London, Six, Expensive}\\
U: Instead, could it be Indian cuisine.\\
\texttt{S: Sure, is there anything else to update?}\\
U: No.\\
\texttt{S: Ok, let me look into some options for you.}\\
\texttt{S: api\_call Indian, London, Six, Expensive}

\subsection{Extended dialog bAbI task 2:}
U: Hi !\\
\texttt{S: Hello, what can I help you with today?}\\
U: I'd like to book a table for six people in an expensive price range with British food in London.\\
\texttt{S: Ok, let me look into some options for you.}\\
\texttt{S: api\_call}\\
\texttt{S (Retrieved Information):}\\
{\texttt{\{The\_Place, british, london,\ldots\}}}\\
{\texttt{The\_Fancy\_Pub, british, london, \ldots\}}}\\
U: Instead, could it be Indian cuisine.\\
\texttt{S: Sure, is there anything else to update?}\\
U: No.\\
\texttt{S: Ok, let me look into some options for you.}\\
\texttt{S: api\_call}\\
\texttt{S (Retrieved Information):}\\
{\texttt{\{Suvai, Indian, london, \ldots\}}}\\
{\texttt{\{Madras\_Masala, Indian, london, \ldots\}}}

\subsection{Original dialog bAbI task 4:}
DB Information given :\\
\textsc{Name}:The\_Place, \space \space \textsc{Cuisine}:British,\\
\textsc{Location}:London, \space \space  \textsc{Price Range}:Expensive,\\
\textsc{Rating}:7, \space \space \textsc{Phone}:The\_Place\_phone,\\ \textsc{Address}:The\_Place\_address, \textsc{Availability}:six\\
U: Hello\\
\texttt{S: Hello, what can I help you with today?}\\
U: I'd like to book a table at The\_Place.\\
\texttt{S: Great, let me do the reservation.}\\
U: May I have the address of the restaurant.\\
\texttt{S: Here it is, The\_Place\_address.}

\subsection{Extended dialog bAbI task 4:}
DB Information given : Entire DB\\
U: Hello\\
\texttt{S: Hello, what can I help you with today?}\\
U: I'd like to book a table at The\_Place.\\
\texttt{S: Great, let me do the reservation.}\\
U: May I have the address of the restaurant.\\
\texttt{S: Here it is, NE\_address.}\\
\texttt{S: The\_Place\_address \textit(retrieved information)}

\section{Multiple-attention based neural retrieval mechanism}
\label{appendix_b}

\begin{figure*}[ht]
\centering
\includegraphics[width=0.9\textwidth]{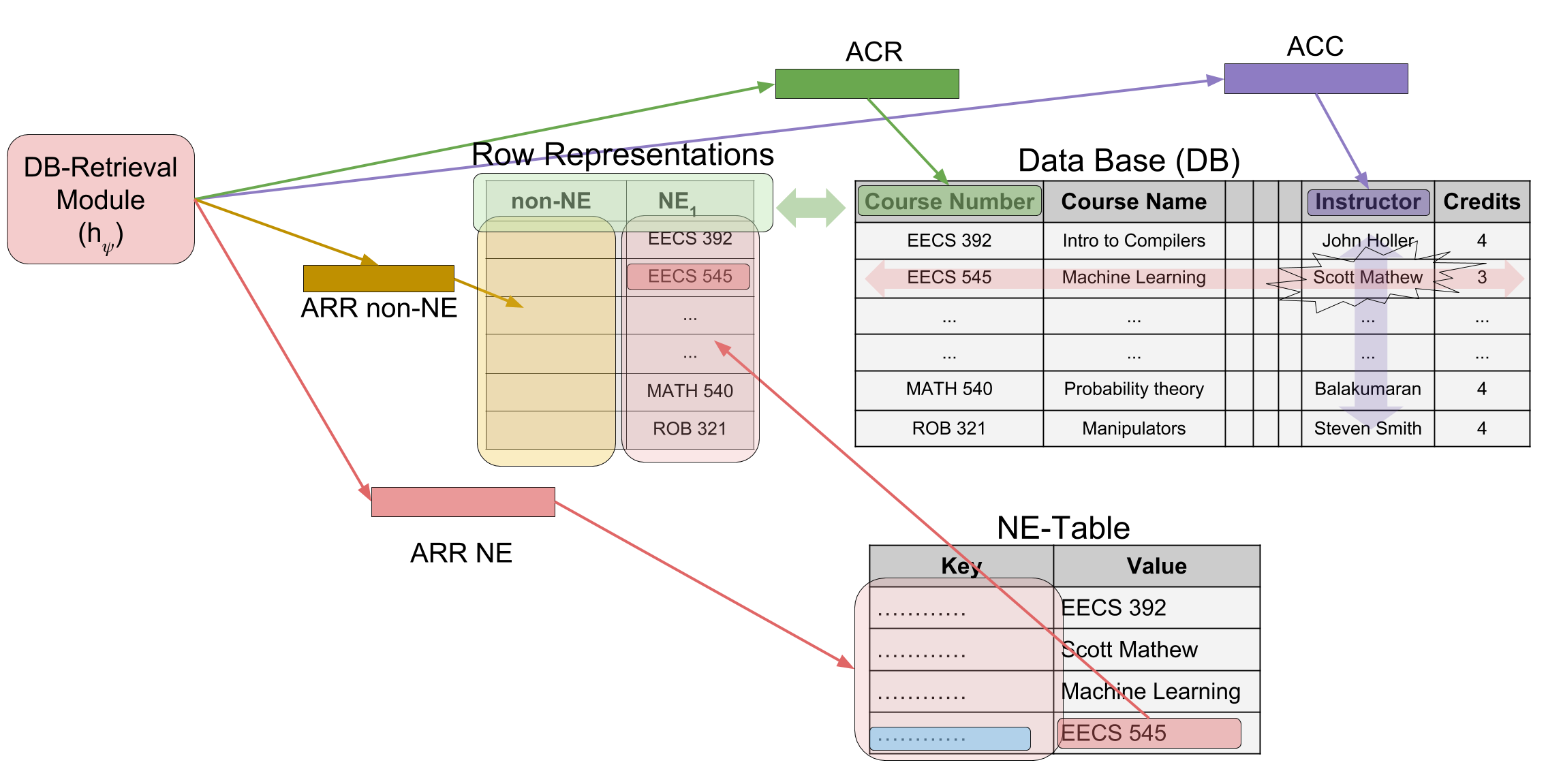}
\caption{Multiple-attention based neural retrieval mechanism. The \textit{DB-Retrieval Module} attends to the relevant rows and columns of the DB by generating attention key embeddings \textit{ACC}, \textit{ACR} and \textit{ARR}.}
\label{fig_att_retrieval}
\end{figure*}

Figure \ref{fig_att_retrieval} shows the schematic of the entire retrieval process. In order to retrieve a particular cell from the table, the system needs to find the correct column and row corresponding to it. 
The \textit{DB-Retrieval Module} ($h_\psi$) does that by generating 3 different attention key embeddings (vectors): Attention over Columns for Columns (\textit{ACC}), Attention over Columns for Rows (\textit{ACR}), Attention over Rows for Rows (\textit{ARR}).

The column(s) that the final retrieved cell(s) belong to, are selected by matching \textit{ACC} key embeddings with the neural embeddings of the column headings (Course Number, Instructor, Credits etc). A separate \textit{ACC} key embedding is generated for every column heading and matched with its embeddings to provide attention scores for all the columns. For the example, \textit{Who teaches EECS545?}, the system would want to retrieve the name of the \textit{Instructor}. Therefore, the \textit{Instructor} column heading alone will have high attention score and be selected. In our experiments, the attention scores are computed through dot products followed by a sigmoid operation, which allows for multiple selections.

Now that the column(s) are chosen, the system has to select row(s), so that it can get the cell(s) it is looking for. Each row in the table contains the values (EECS545, Machine Learning, Scott Mathew etc) of several attributes (Course Number, Course Name, Instructor etc). But we want to assign attention scores to the rows based on particular attributes that are of interest (\textit{Course Number} in this example). The column/attribute headings that the system has to attend to for selecting these relevant attributes are obtained by matching \textit{ACR} (Attention over Columns for Rows) key embeddings with the neural embeddings of the different column headings. 

The last step in the database retrieval process is to select the relevant rows using the \textit{ARR} (Attention over Rows for Rows) key embedding. \textit{ARR} is split into two parts \textit{ARR NE} and \textit{ARR non-NE}. In a general scenario, \textit{ACR} can select multiple columns to represent the rows. For each selected column that is a NE column, a separate NE-value is retrieved from the \textit{NE-Table} using a separate \textit{ARR NE} embedding for each of them. These NE values are used to do exact match search along the corresponding columns (in the NE row representations) to select the matching rows. For the non-NE columns that are selected by \textit{ACR}, their neural embeddings are combined together along each row to get a fixed vector representation for each row in the DB (weighted sum of their embeddings, weighted by the corresponding column attention scores). \textit{ARR non-NE} is then used to match these representations for selecting rows. The intersection of the rows selected in the NE row representations and the non-NE row representations is the final set of selected rows.

In short, the dialog system can use neural embedding matching for non-NEs, exact value matching for NEs and therefore a combination of both to decide which rows to attend to. Depending on the number of columns and rows we match with, we select zero, one or more output cells. For our running example, \textit{ARR NE} is used to match with the keys in the \textit{NE-Table} to select the row corresponding to \textit{EECS 545} and the value \textit{EECS 545} is returned to do an exact match over the NE row representations (represented by the course number values). This gives us the row corresponding to \textit{EECS 545} and hence the cell \textit{Scott Mathew}. We could use our \textit{NE-Table} idea with potentially many types of neural retrieval mechanisms to retrieve information from the DB. The multiple-attention based retrieval mechanism, described above, is only one such possible mechanism. 

{ 
\renewcommand{\arraystretch}{1.5}
\begin{table*}[t]
\centering
{
\scriptsize
\begin{tabular}{c|c|c|c|c|c|c|c|c}
{Task}&{Model} & {ACR} & {ARR non-NE} & {ARR NE} & {DB-Retrieval} & {Per-response} & {Per-Dialog} & {Per-Dialog + DB-Retrieval}\\ \hline
\multirow{2}{*}{Task 1} & W/O-NE-Table & 100 (100) & 9.0 (6.9) & - & 10.2 (7) & 100 (98.2) & 100 (90.3) & 10.2 (6.7)\\ \cline{2-9}
&With-NE-Table & 99.4 (98.1) & 96.9 (96.7) & 100,100 (100,100) & 98.5 (99.0) & 99.8 (99.8) & 98.8 (99) & 97.3 (98.0) \\ \hline 
\multirow{2}{*}{Task 2} & W/O-NE-Table & 100 (100) & 8.6 (7.6) & - & 0.8 (1.0) & 100 (100) & 100 (100) & 0.0 (0.1)\\ \cline{2-9}
 &With-NE-Table & 100 (100) & 99.1 (99.8) & 100,100 (100,100) & 99.6 (99.8) & 100 (100) & 100 (100) & 99.2 (99.7) \\ \hline 
\end{tabular}
\caption{Results for extended dialog bAbI task 1 and 2. Accuracy \% for Test and Test-OOV (given in parenthesis). ARR non-NE columns are price and number of people. ARR NE columns are cuisine and location. DB-Retrieval \%: Retrieval accuracy for rows (task 1,2) and a particular cell (task 4). Per-Dialog \%: Percentage of dialogs where every dialog response is correct. Per-Dialog + DB-Retrieval \%: Percentage of dialogs where every dialog response and information from DB retrieval are correct.}
\label{table_babi_task1_2_extended}
}
\end{table*}
\normalsize{}
}

{
\renewcommand{\arraystretch}{1.5}
\begin{table*}[h]
\centering
{\scriptsize
\begin{tabular}{c|c|c|c|c|c|c|c|c}
{Model} & {ACR} & {ACC} & {ARR non-NE} & {ARR NE} & {DB-Retrieval} & {Per-response} & {Per-Dialog} & {Per-Dialog + DB-Retrieval}\\ \hline
W/O-NE-Table & 100 (100) & 100 (100) & 0.0 (0.0) & - & 0.0 (0.0) & 100 (100) & 100 (100) & 0.0 (0.0)\\ \hline
With-NE-Table & 100 (100) & 100 (100) & - & 100 (100) & 100 (100) & 100 (100) & 100 (100) & 100 (100) \\ \hline 
\end{tabular}
\caption{Results for extended dialog bAbI task 4. Accuracies in \% for Test and Test Out-Of-Vocabulary (given in parenthesis). DB-Retrieval \%: Retrieval accuracy for rows (task 1,2) and a particular cell (task 4). Per-Dialog \%: Percentage of dialogs where every dialog response is correct. Per-Dialog + DB-Retrieval \%: Percentage of dialogs where every dialog response and information from DB retrieval are correct.}
\label{table_babi_task4_extended}
}
\end{table*}
\normalsize{}
}

\renewcommand{\arraystretch}{1.5}
\begin{table*}[h]
\centering
\scriptsize
\begin{tabular}{p{2.0cm}|p{4.5cm}|p{2.0cm}|p{1.6cm}|p{1.6cm}|p{1.4cm}}
Task & Model & Evaluation & Task 1 & Task 2 & Task 4 \\ \hline
Original bAbI tasks & Baseline(MemN2N + match-type + RNN-encoding) & Per-Dialog & 100 (100) & 99.9 (50.6) & 100 (100)\\ \hline
Extended bAbI tasks & With-NE-Table & Per-Dialog + DB-Retrieval & 97.3 (98.0) & 99.2 (99.7) & 100 (100) \\ \hline
\end{tabular}
\caption{Performance comparison of our model in the extended dialog bAbI tasks, with a baseline model in the original bAbI tasks. Accuracies in \% for Test and Test Out-Of-Vocabulary (given in parenthesis).}
\label{table_babi_original}
\end{table*}
\normalsize{}

\section{Goal oriented dialog tasks: extended results }
\label{appendix_c}

\subsection{Extended results for tasks 1 and 2}
The detailed results for task 1 and task 2 are shown in Table \ref{table_babi_task1_2_extended}.

\textit{With-NE-Table}: For issuing an \textit{api\_call} in tasks 1 and 2, four argument values are required - cuisine, location, price range and number of people. We consider cuisine and location to be NEs. So whenever cuisine and location names occur in the dialog, a NE key is generated on the fly and is stored in the \textit{NE-Table} along with the NE values. 

\begin{itemize}
\itemsep0em
    \item ACC: For tasks 1 and 2, ACC is not required as we are interested in retrieving rows.
    \item ACR: ACR is used to select the columns required to represent the rows. These are four columns - NE columns (cuisine and location) and non-NE columns (price range and number of people)
    \item ARR-non-NE: Each row in the DB is represented by weighted vector (embedding) sum of its price range and number of people (embeddings). The model returns the relevant rows using attention on the non-NE columns embeddings. 
    \item ARR-NE: The model attends over the \textit{NE-Table} by matching (dot product) its generated key with the keys present in the \textit{NE-Table} to retrieve NE values. The selected NE values are then matched (exact-match) with cuisine and location values in DB to retrieve the relevant rows. 
    \item The final retrieved rows are the intersection of the rows selected by ARR-non-NE and ARR-NE.
\end{itemize}

\textit{W/O-NE-Table}: \textit{ACR} is used to attend to the four relevant columns. However, each row is represented by the combined neural embedding representation of all four attribute values, cuisine, location, price range and number of people. \textit{ARR non-NE} is used to retrieve the relevant rows.

From Table \ref{table_babi_task1_2_extended}, we can see that both the models perform well in selecting the relevant columns, but the model \textit{W/O-NE-Table} performs poorly in retrieving the rows, while \textit{With-NE-Table} performs very well. This results in \textit{With-NE-Table} model achieving close to 100\% accuracy in DB retrieval while \textit{W/O-NE-Table} performs poorly.

This is because, in the \textit{With-NE-Table} model, the task of retrieving rows is split into two simpler tasks. The NEs are chosen from the \textit{NE-Table}, and then exact matching is used (which helps in handling OOV-NEs as well). The non-NEs, price range and number of people, have limited set of possible values (low, moderate or expensive for price range and 2,4,6 or 8 for number of people respectively). This allows the system to learn good neural embeddings for them and hence have high accuracy in \textit{ARR non-NE}. Whereas in \textit{W/O-NE-Table} model, \textit{ARR non-NE} involves the neural representations of cuisine and location values as well, where a particular location and cuisine value will occur only a few number of times in the training dataset. In addition to that, new cuisine and location values can occur during the test time (Test OOV dataset, performance shown in parenthesis).

For the dialog part (which does not involve the DB retrieval aspect) of extended tasks 1 and 2, the system utterances do not have any NEs in them. However, the user utterances contain NEs (cuisine and location that the user is interested in) and so the system has to understand them in order to select the right system utterance. The accuracy in performing the dialog (by selecting responses from candidate set) is similar for both the models on the normal test set. However, in the OOV-test set, for task 1, where the system has to maintain the dialog state to track which attribute values have not been provided by the user yet, \textit{W/O-NE-Table} model seems to get affected, while the \textit{With-NE-Table} model is robust to that. While \textit{W/O-NE-Table} gets a Per-Dialog accuracy of 90.3\% in the OOV-test set, \textit{With-NE-Table} is able to get 99\%.\\

\subsection{Extended results for task 4}
Detailed results for task 4 are shown in Table \ref{table_babi_task4_extended}.

\textit{With-NE-Table}: In task 4, the user tells the system the restaurant in which he/she wants to book a table. The restaurant name, which is a NE, is stored in the \textit{NE-Table} along with it's generated key. When the user asks for information about the restaurant such as, phone number, the NE restaurant name stored in the \textit{NE-Table} is selected and used for retrieving its corresponding phone number from the DB. For this particular case, \textit{ACC} attends over the column \textit{Phone} and \textit{ACR} attends over \textit{Restaurant Name}. Since the column selected by \textit{ACR}  is a NE column, the NE value (here the actual restaurant name given by the user) is retrieved using \textit{ARR NE} from the \textit{NE-Table}. The retrieved NE value is used to do an exact match over the DB column selected by \textit{ACR} to select the rows. The cell that intersects the selected row and the column selected by \textit{ACC} is returned as the retrieved information and used to replace the NE type tag in the output response.

\textit{W/O-NE-Table}: Here, all input words (including NEs) are part of the vocabulary and for NEs, their embedding given to the sentence encoder is the sum of the NE word embedding and the embedding associated with its NE-type. The candidate response retrieval (dialog) is same as the above model and the column attentions are also similar. However, the models differ with respect to attention over rows. Since NEs are not treated special here, attention over rows happens through \textit{ARR non-NE}. For this task, when \textit{ACR} is selected correctly (restaurant name), each row will be represented by the neural embedding representation of its restaurant names. \textit{ARR non-NE} generates a key to match these neural embeddings to attend to the row corresponding to the restaurant name mentioned by the user.

\section{Comparison with original dialog bAbI tasks}
We choose the best model (MemN2N + match-type features) from \cite{bordes2016learning-memN2N} (they use match-type features for dealing with entities) and update the baseline model by using RNN encoding for sentences (similar to \textit{With-NE-Table}). Note that we achieve higher accuracy for our updated baseline model for original bAbI tasks than reported in \cite{bordes2016learning-memN2N}, which we attribute to the use of RNN for encoding sentences (they use BoW encoding). 

For match-type features, \cite{bordes2016learning-memN2N} add special words (\textit{R\_CUISINE}, \textit{R\_PHONE} etc.), for each KB entity type (cuisine, phone, etc.) to the vocabulary. The special word (e.g. \textit{R\_CUISINE}) is added to a candidate if a cuisine (e.g. Italian) appears in both dialog and the candidate. 
For each type, the corresponding type word is added to the candidate representation if a word is found that appears 1) as a KB entity of that type, 2) in the candidate, and 3) in the input or memory.
For example, for a task 4 dialog with restaurant information about \textit{RES1}, \emph{only} one candidate \textit{"here it is RES1\_phone"} will be modified to \textit{"here it is RES1\_phone R\_PHONE"}. 
Now, if the user query is for the restaurant's phone number, using match-type features essentially reduces the output search space for the model and allows it to attend to specific candidates better. Hence, match-type features can only work in a retrieval setting and will not work in a generative setting. Our \textit{With-NE-Table} model will work in both retrieval and generative settings.

Table \ref{table_babi_original} compares the performance of the \textit{With-NE-Table} model in the extended bAbI tasks with that of a baseline method on the original bAbI tasks. Note that extended dialog bAbI tasks require the dialog system to do strictly more work compared to the original dialog bAbI tasks. Though not a strictly fair comparison for our model, we observe that the performance of our \textit{With-NE-Table} model in extended bAbI tasks is as good as the performance of updated baseline model in original bAbI tasks. In addition to that, for bAbI task 2 OOV test set, \textit{With-NE-Table} model performance in the extended bAbI task, is actually much higher compared to the baseline model on the original bAbI task (99.7\% vs 50.6\%).

\end{document}